\documentclass[letterpaper, 10 pt, conference]{ieeeconf}  

\IEEEoverridecommandlockouts                              
\title{\LARGE \bf
A Concise Introduction to Reinforcement Learning in Robotics
}

\author{ \parbox{2.2 in}{\centering Akash Nagaraj\\
         Department of Computer Science\\
         PES University\\
         Bangalore, India\\
         {\tt akashn1897@gmail.com}}
         \parbox{2.2 in}{ \centering Mukund Sood\\
         Department of Computer Science \\
         PES University\\
         Bangalore, India\\
         {\tt mukundsood2013@gmail.com}}
          \parbox{2.2 in}{ \centering Bhagya M Patil\\
         Department of Computer Science \\
         PES University\\
         Bangalore, India\\
         {\tt bhagyampatil@pes.edu}}
}
\usepackage{graphicx}
\usepackage{xcolor}
\usepackage{color}
\usepackage{relsize}
\usepackage{dblfloatfix}
\begin{document}

\maketitle
\thispagestyle{empty}
\pagestyle{empty}

\begin{abstract}
One of the biggest hurdles robotics faces is the facet of sophisticated and hard-to-engineer behaviors. Reinforcement learning offers a set of tools, and a framework to address this problem. In parallel, the misgivings of robotics offer a solid testing ground and evaluation metric for advancements in reinforcement learning. The two disciplines go hand-in-hand, much like the fields of Mathematics and Physics. By means of this survey paper, we aim to invigorate links between the research communities of the two disciplines by focusing on the work done in reinforcement learning for locomotive and control aspects of robotics. Additionally, we aim to highlight not only the notable successes but also the key challenges of the application of Reinforcement Learning in Robotics.\\
This paper aims to serve as a reference guide for researchers in reinforcement learning applied to the field of robotics. The literature survey is at a fairly introductory level, aimed at aspiring researchers. Appropriately, we have covered the most essential concepts required for research in the field of reinforcement learning, with robotics in mind. Through a thorough analysis of this problem, we are able to manifest how reinforcement learning could be applied profitably, and also focus on open-ended questions, as well as the potential for future research.

\end{abstract}

\textbf{\textit{ Keywords- reinforcement learning in robotics, robotic learning, learning locomotion, learning actions.}}
\section{INTRODUCTION}

Reinforcement Learning (RL) is an important type of Machine Learning where an \textit{agent} learns how to behave in a particular environment by performing actions and observing the results of these actions. Based on these results the \textit{agent} is assigned a \textit{reward}. \par

Numerous challenging problems have seen recent success when addressed with model-free Reinforcement Learning algorithms. These have proven to make very effective problems to evaluate algorithms because of the fact that they are entirely observable, manipulable and have the ability to run quickly and simultaneously gather vast amounts of experience for the agent. They have the potential for not only structural, but also perpetual similarities to real-world problems. \par

Now, tasks pertaining to the real-world serve as a target for Reinforcement Learning algorithms, especially in the field of robotics which bear both control problems and difficult perception. Recent notable contributions include motor control directly from pixels for manipulation, continuous control for simulated robots, and simulation-to-real transfer, also for manipulation. \par

Recent work has also focused on addressing the task of navigation in 3D environments in simulation, such as maze navigation using imagery from a perspective view. In the field of robotics, agents have also been trained to navigate with the help a number of different sensors. Visual sensing is seen as an established direction. It includes work on fine-tuning in the real world as well as target-driven image-based navigation in small grid worlds with a simulation. Structure-based sensors are also used because of their ability to exhibit minor difference between simulation and reality than seen in visual sensing. \par

To understand the need for a reward, we will use an analogy. Assume there is a child in a room with a fireplace. As the child moves closer to the fireplace it feels the warmth, and likes it. This would be considered a positive reward. However, if the child were to go and touch the fire, he would burn himself. This would be considered a negative reward. Therefore, we can say that the child has learned from the observations obtained from the actions he took. This is the fundamental need for a \textit{reward} system in Reinforcement Learning. \par

The way a Reinforcement Learning task is defined is as follows:
\begin{itemize}
    \item An agent receives \textbf{state $S_0$} from the environment
    \item Based on the \textbf{state $S_0$}, the agent will take an \textbf{action A0}
    \item The environment then transitions to a new \textbf{state $S_1$}
    \item The environment returns some \textbf{reward $R_0$} to the agent
\end{itemize}
This way, the Reinforcement Learning model returns a sequence of \textbf{state, action, and reward} with the goal of the agent to maximize the expected cumulative reward. \par

This whole process can also be called a \textit{Markov Decision Process} (MDP). An MDP is established using the Markov Property, which states that \textit{given the present, the future is independent of the past}. In mathematical terms, a state St has the Markov Property, if and only if: 
\begin{center}
    $P(S_{t+1}|S_t) = P(S_{t+1}|S_1, ..., S_t)$
\end{center}
Formally, and MDP is used to describe the environment for RL, where the environment is fully observable. Almost all RL problems can be formalized as MDPs. 

Mathematically, we can define the cumulative reward stated above as a Markov Reward Process by the equations:
\begin{center}
$ G_t = \mathlarger{\sum_{k=0}^{\infty} \gamma^k R_{t+k+1}}$; where $\gamma \epsilon [0,1)$  \\
\bigbreak
$R_{t+1} + \gamma R_{t+2} + \gamma^2 R_{t+3}...$
\end{center}

It is logical to assume while calculating the cumulative reward, that earlier rewards have a higher probability of happening, as they are more predictable than the long-term future reward. Hence, it is important to discount rewards, and this is done by introducing a \textit{discount rate gamma} whose value must be between 0 and 1. Another need for the discount rate is that it is mathematically convenient and guarantees that the algorithm will converge, avoiding infinite returns in loopy Markov Processes. As seen in the equation, the addition of gamma can be interpreted in two ways:
\begin{itemize}
    \item A larger gamma results in a smaller discount, implying the agent is more worried about the long-term reward
    \item A smaller gamma results in a larger discount, implying the agent is more worried about the short-term reward. 
\end{itemize}

 \begin{figure*}[!b]
    \begin{center}
        \includegraphics[width=0.9\textwidth]{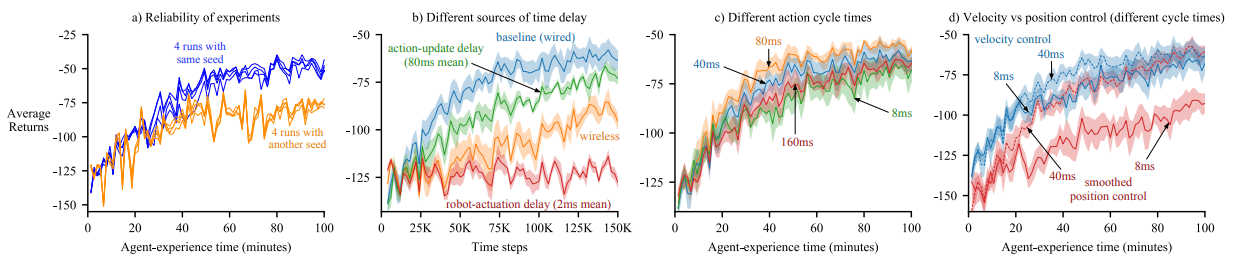}
        \caption{Learning performance in different UR5 Reacher setups. a) Our baseline setup. b) Artificial delays. c) Action cycle times resulted in worse learning performance. d) Velocity control.}
        \label{fig:F43}
    \end{center}
\end{figure*}
\section{Setting Up A Reinforcement Learning Task with a Real-World Robot}

The aim of this work is to address the question of how to set up a real-world robotic task so that an off-the-shelf implementation of a standard RL method can perform effectively and reliably. Rupam et al. designed a \textit{Reacher} task with the UR5 Robot (called \textit{UR5 Reacher}). The task of the agent is to learn to reach arbitrary target positions with low-level actuations of a robot arm using a reinforcement learning approach. \par

Through this paper, they try to highlight the difference between simulations of RL tasks and their real-world counterparts. They then go on to discuss the steps and elements of setting up real-world RL tasks including a medium of data transmission, concurrency, ordering and delays of computation, low-level actuation types, and frequencies of operating them. By varying these elements, they study and report their individual contributions and the difficulty with robot learning. They show that by accounting for the effects of these variations and addressing them with care during the set-up of the robot for the RL task, it is possible to achieve not only effective performance, but also repeat-ability of learning from scratch in a highly reliable manner even when the repeats are run for hours on different times using different physical robots. \par

Below is a summary of each of the elements involved in a real-world setup and the effect that they have on each experiment:

\subsection{Concurrency, ordering, and delays of computation}
During simulations of the same task, it is natural to perform agent and environment-related computations synchronously, which may not be desirable in real-world tasks. Simulated tasks can comply with the Markov Decision Process (MDP) framework where time does not advance between observing and acting, which is not the case in the real-world. In the real-world time marches on during each agent and environment-related computation. Hence, the agent is always operating on delayed sensorimeter information. This can lead to misplaced synchronization and ordering of computations, which may result in a more difficult learning problem and reduced potential for responsive control. Therefore, a design objective should be to manage and minimize these delays. \par
In \textit{UR5 Reacher}, the computational steps by distributing them into two asynchronous processes: the \textit{robot communication process} and the \textit{reinforcement learning (RL) process}.\par
The robot communication process is a device driver which collects the sensorimotor data in a \textit{sensor thread} and send the actuation commands in a separate \textit{actuator thread}. The RL process however, contains an \textit{environment thread} that checks spatial boundaries, computed the observation vector and the reward function that is based on the sensorimotor packets and further updates the actuation command for the actuator thread. It also contains an \textit{agent thread} to define task time steps and determining the action cycle time. Hence,  splitting the RL process into two threads allows checking safety constraints faster than, and concurrently with action updates. 

\subsection{Medium of data transmission}
 Here Reference [1] highlights the natural consideration of a scientist to pair a mobile robot with limited on-board computing power with a more computationally powerful base station via a WiFi or Bluetooth module as opposed to the standard USB or Ethernet option available. This however, comes with its own drawbacks. \par
 Often, WiFi brings in variability in the inter-arrival time of streamed packets. In the task setup of the Reference [1], they initially set-up a baseline model using an Ethernet connection for the robot communication process, before switching to communication over a TCP/IP connection.
 
 \subsection{Action cycle time}
 This is the time between two subsequent action updates by the agent's policy, it is also commonly called the time-step duration. \par
 Reference [1] highlight how choosing a cycle time for the task at hand is not obvious, and the available literature lacks guidelines for a task set-up. A shorter cycle time may include superior policies with finer control, however, if changes in subsequent observation vectors with too short cycle times are not perceptible to the agent, it results in a learning problem that is very difficult and near impossible. At the other end, a longer cycle time will limit the set of possible policies and the control one has, however, it may make the learning problem easier. \par
 Hence, it is clear that a trade-off must be made during conduction of the real-world task to obtain an optimal output.
 
\subsection{The action space: position v/s velocity control}
When the learning process is in its infant stages, the initial policies learned by the robot resulted in sequences of angular position commands that caused violent, abrupt movements and emergency stops. Hence, initially it is possible to limit the speeds the robot can take, thus avoiding the sudden jerks. \par
Modulation of position and velocity commands is an important aspect.

\section{Solving Sparse Reward Tasks from Scratch}

Reference [2] defines the problem to be for a robot to perform tasks such as opening a box and placing a block inside it, stacking blocks on top of each other etc. We must note that while defining the final rewards are easy, such as assigning a reward of 1 when a block is placed in a box successfully, or stacking one block on another, the complete action has several intermediate steps that must also be given rewards. Hence, this becomes a problem with sparse rewards, and the learning process becomes harder. \par

A new algorithm is defined for robots to learn such problems through RL, called Scheduled Auxiliary Control (SAC-X). Their approach however, is based on four main principles, as described in their paper. Those being: 
\begin{itemize}
    \item Every state-action pair is paired with a vector of rewards (which are of two types - internal auxiliary rewards and external rewards)
    \item Each reward entry has an assigned policy, called an intention, that is trained to maximize its corresponding cumulative reward. 
    \item A high-level scheduler will select and execute the individual intentions with the goal of improving the performance of the agent on the external tasks. 
    \item Learning is performed asynchronously from policy execution and the experience between intentions is shared so as to use the information effectively. 
\end{itemize}

\subsection{Scheduled Auxiliary Control (SAC-X)}
The algorithm is a hierarchical approach to reinforcement learning for learning from sparse rewards. The goal defined is to both
\begin{itemize}
    \item train all auxiliary intention policies as well as the main task policies to achieve their respective goals.
    \item utilize all intentions for fast exploration in the main sparse-reward Markov Decision Process. 
\end{itemize}
The second point was achieved by using a hierarchical objective for policy training that decomposes into two parts
\begin{itemize}
    \item Learning the intentions - the first part of the hierarchical objective is given by joint policy improvement objective for all intentions (auxiliary and external)
    \item Learning the scheduler - the second part of the hierarchical objective deals with learning a scheduler that sequences intention-policies. 
\end{itemize}
\begin{figure*}[!b]
    \begin{center}
        \includegraphics[width=0.9\textwidth]{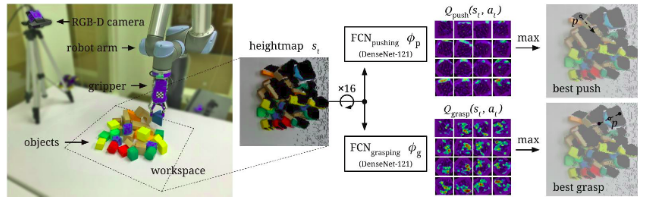}
        \caption{Overview of the system and Q-learning formulation}
        \label{fig:F43}
    \end{center}
\end{figure*}
\subsection{Conclusion}
The new algorithm described in Reference [2] is run on a variety of experiments. They compare the performance of their novel method to the previous ones already defined and in existence. [6]  \par
The experiments that they carried out were predominantly variations of stacking experiments (stacking a smaller object on top of a larger object and vice-versa, stacking irregularly shaped objects on top of other objects, clean-up tasks etc.). They also show that they can learn certain tasks that were unable to be learned by these previous algorithms. \par
They prove that by adding a scheduler, which based on the current state and information from the external world, maps out a sequence of events that will give the highest reward, the robot is able to learn better and most importantly, more efficiently. \par

\section{Learning Synergies between Prehensile and Non-Prehensile Actions}
Prehensile and non-prehensile actions are an integral part of robotics. Prehensile and non-prehensile go hand-in-hand, for instance, pushing (prehensile) can help rearrange cluttered objects in order to provide space for arms and fingers of the robot to facilitate a successful grasp. Likewise, grasping (non-prehensile) is useful to displace objects to facilitate collision-free as well as precise pushing motions. \par
Another approach to this is the hard-coding, which exploits domain-specific knowledge, however, this severely limits the various types of synergistic behaviors between prehensile and non-prehensile actions that can be performed. It must also be duly noted that pushing actions, are considered useful only if, in time, they facilitate grasping actions. \par
Reference [3] which lies in the intersecting domains of machine learning, robotic manipulation, and computer vision, is one that focuses on learning of the synergy between pushing and grasping with the help of self-supervised Deep Reinforcement Learning. The primary contribution of [3] is that it shines light on a completely new perspective to bridging the gap between data-driven prehensile and non-prehensile manipulation.
Reference [3] has an architecture that comprises of two, fully convolutional networks that are used to map visual observations of a camera to robotic actions; prehensile and non-prehensile. The first network infers the usefulness of a grasp in a dense sampling (done pixel-wise) end-effector orientations and locations, while the second fully convolutional network does the exact same, but in this case for pushing.
FCNs $\phi_p$ (pushing) and $\phi_g$ (grasping) share the same network architecture: two parallel 121-layer DenseNets pre-trained on ImageNet[7], proceeded by channel-wise concatenation and two 1x1 convolutional layers interleaved using nonlinear activation functions (ReLU), and spatial batch normalization, and finally bi-linearly up-sampled. A small learning rate is prescribed to the network at $10^{-5}$ and gradients are back-propagated through the network after each executed action.\par
The system is trained jointly in a Q-learning [10] framework and are entirely self-supervised by trial and error, and rewards are provided based on successful grasps. The policies are trained end-to-end using a deep network that accepts visual observations and outputs the expected return (in the form of Q values) for potential pushing and grasping actions, given the visual observation. In all, 32 pixel-wise maps of Q values (16 for prehensile actions (pushes) in different directions, and 16 for non-prehensile actions (grasps) at different orientations) are considered. The joint policy is used to choose the action with the maximum Q value in order to maximize the chance of successful current/future grasps. This system is also proven to generalize very well to novel objects. \par
The Markov decision process representation of the entire system is as follows: \textit{at any given state $s_t$ at time $t$, the agent (i.e. robot) chooses and executes an action at according to a policy $\pi(s_t)$, then transitions to a new state $s_{t+1}$ and receives an immediate corresponding reward $R_{at}$ ($s_t$, $s_{t+1}$). The goal of our robotic reinforcement learning problem is to arrive at an optimal policy $\pi^\ast$ that ensures maximization of the expected sum of future rewards.}

$R_g(s_t, s_{t+1}) = 1$ is awarded in the event of a successful grasp, which is computed by calculating the threshold of the antipodal distances between the gripper fingers following a grasp attempt. Similarly, $R_p(s_t, s_{t+1}) = 0.5$ for pushes that cause detectable changes to the environment. \par

The system models each state $s_t$ as an RGB-D height-map image representation of the scene at a particular time $t$. For the purpose of computation of this height-map, RGB-D images are captured with the help of a fixed-mount camera. These images are then projected onto a 3D point cloud, and back-projected, orthographically against in the direction of gravity (upwards) to construct a height-map image representation with both height-from-bottom (D) as well as colour (RGB) channels. \par

\section{Locomotive Behaviours in Rich Environments}

One of the important and immanent aspects of robotics is locomotion. Locomotive movements in robots are notoriously known to be behaviours that are very sensitive to the choice of reward. Agents can be taught to perform various actions such as running, crouching, jumping and turning, as needed by the environment in the absence of explicit reward-based guidance. \par 
Reference [4] is a paper that shows great promise in this area of research by building upon robust, pre-existing policy gradient algorithms, for instance: Trust Region Policy Optimization (TRPO) [8] and Proximal Policy Optimization (PPO) [9]. These algorithms operate by bounding each parameter update to a \textit{trust region}, thereby ensuring stability by restricting the amount by which a particular update can change the policy. Additionally, similar to the A4C algorithm [11] and its related approaches, computation is distributed over numerous distributed, parallel instances of both the environment, and the agent. This distributed implementation of Proximal Policy Optimization (PPO) shows a significant improvement over TRPO with respect to wall clock time, that being said, we still see minor differences when it comes to robustness of the algorithms. It also shows an improved performance over many existing implementations of the A3C algorithms, using the same number of workers, with continuous actions provided. \par

In every iteration of TRPO given the current parameters (denoted by $\theta_{old}$), a comparatively large batch of data is collected, which aims to optimize the surrogate loss. The constraint is placed on the magnitude that the policy is allowed to change, and we express this in terms of the Kullback-Leibler divergence. \par

Distributed PPO is very similar to TRPO in terms of performance. DPPO scales very well proportional to the number of workers used, possible resulting in significant reduction of wall clock time required. Since DPPO is a fully gradient based algorithm, therefore, as shown in the Memory Reacher Task [4], it can also be used with Recurrent Networks. \par

The experimental setup [4] comprises of a physical simulation environment similar to a platform game, implemented in Mujoco. \par

Three different torque-controlled bodies are considered: \textcolor{white}{\tiny{"}}
\begin{itemize}
\item \textbf{Planar walker:} a simple walking body with 9 Degrees of Freedom (DoF), and 6 actuated joints constrained to the plane. 
\item \textbf{Quadruped:} a simple three-dimensional quadrupedal body with 12 DoF and 8 actuated joints. 
\item \textbf{Humanoid:} a three-dimensional humanoid with 21 actuated dimensions and 28 DoF.  \textcolor{white}{\tiny{"}}
\end{itemize} \par

The rewards for all tasks are kept consistent and simple across all terrains. The reward comprises of a main component, which is in proportion to the velocity along the x-axis, plus a small term penalizing torques. This encourages the agent to make progress in the forward direction along the track.
The agent receives two collections of observations: 
\begin{enumerate}
    \item a collection of egocentric, "proprioceptive" features containing joint angles as well as angular velocities. In cases of the Humanoid and Quadraped, they also contain the readings for a gyroscope, accelerometer, and a velocimeter that are placed at the torso, in addition to contact sensors attached to the feet and legs.
    \item a collection of “exteroceptive” features which contains information that is task-relevant, such as the profile of the terrain ahead, as well as the position of the agent with respect to the center of the track.
\end{enumerate}

It must also be noted that the walker also occasionally overcomes obstacles by chance. The probability associated with this change event is minute when the height of the obstacles are very high.

\section{One-Shot Reinforcement Learning for Navigation}

Challenging problems have been solved by learning from an agent's extensive interaction with the environment by using model-free reinforcement learning algorithms. Reference [5] present a new method to teach an agent to learn to navigate to a fixed destination or goal in a set environment. A primary priority of any robotic learning system is to be able to learn a behaviour or task while still managing to  minimize the interaction needed with the environment. {\color{white}{\tiny{"}}} \par

In order to learn to navigate towards a fixed destination or goal in a known, real-world environment reliably, Reference [5] proposes a method that performs an interactive replay of a single traversal of the said environment. To accomplish this, the learning model uses pre-trained visual features, and also augments the training set with various stochastic observations to demonstrate zero-shot transfer to variations in the real-world environment which are unseen during training. \par

In model-free Reinforcement Learning, the policy is implemented with the help of a function approximator, a deep neural network is commonly used for this application. Customarily, the system is usually trained either to approximate the probability distribution over the actions of the agent directly, ensuring the maximum expected reward (as in policy search) or to map actions and observations to the expected future reward (as in Q-learning). Coming to value-based Reinforcement Learning methods such as Q-learning, the policy consists of a value estimate (the Q function) being used to choose actions with the highest return. In comparison, in the case of policy search methods, the policy is parameterized directly and the parameters of the policy are updated in the direction of positive outcomes using a value function. In the case of [5], a value-based method known as bootstrapped Q-learning is used.\textcolor{white}{\tiny{"}} \textcolor{white}{\tiny{"}}\par 

Interactive Replay is a concept in which an agent uses a single traversal of the environment to memorize a rough world model. This provides the agent with an opportunity to interact with the model and generate s significant number of diverse trajectories. The purpose of this is to learn how to navigate, while also minimizing the amount of real-world data required by the agent. Using interactive replay, the agent learns to navigate through the environment as a single trajectory through a real environment comprises of a virtual environment. In addition, a validation environment is also constructed using a second pass of the environment however with different environmental conditions, which is approximately aligned with the training environment. \par

First, the agent executes a complete traversal of the environment, and records all the sensor data during the traversal. Second, the recorded data is arranged in the form of a topological map of the environment. In the experimental setup in [5], minimal human interaction is required to align the data to a topological map. In the map, a human is shown the approximate poses of the agent in 2-Dimensional space, loops are closed, and duplicated regions of space are removed. Generally, both phases can be accomplished autonomously using only Simultaneous Localization and Mapping (SLAM) techniques without human intervention. A pose graph consisting of sensory snapshots taken at intervals of $\Delta_{pos}$ meters is created by dicretizing the environmental space. \par

Collection of the validation environment is simpler when compared to the collection of a training environment. The agent conducts a second traversal of the environment, which need not necessarily be in the order as the first traversal. Accordingly, the training environment is used with an approximate localization method to associate the training environment to nodes in the pose graph, with the closest matching pose. In [5], laser-based localization is used as the approximate localization system. This is coupled with a $360^{\circ}$ camera used to align the image's orientation, but pose correspondences could as well be annotated by hand in the case when a localization system is unavailable. \textcolor{white}{\tiny{"}}\par

One of the consequences of training your model on a small set of real-world experience is that over-fitting to the constrained environment. Over-fitting could also manifest in the learning of visual features for environment identification. Most of the current methods in model-free RL aim at learning feature extractors from absolutely nothing. They use just one traversal of a small environment, and this proves an extremely limited variety of visual structure. The lack of diversity in the training environment can be compensated by using a visual encoder network trained on a very large vision dataset. \par

Another solution for the issue of limited data is data augmentation. It i a proven and widely used technique in machine learning. [5] exploits this technique by augmenting the training environment with stochastic observations to aid in learning. These techniques have been established to compensate for the lack of extensive training samples, in which several random transformations are applied to the training samples to facilitate the improvement of the diversity of the training data. \par

Model-free Reinforcement Learning algorithms involve a double dueling $n$-step Q-learning framework established using $N_Q$ parallel Q-Function heads. Each individual head is a Deep Neural Network in it's own-self, with recurrent layers and shared visual encoders, but with its own Q-function output layers.

Reference [5] trains the bootstrapped Q-Network to minimize the traditional Q-learning loss function:
\begin{center} 
    $L = \mathlarger{\Sigma_{batch}}Q(f(x_t)) - R_t)^2$
\end{center}
where, $R_t$ is the sum of exponentially-discounted future rewards. The ensemble nature of bootstrapped Q-learning is that it's ensemble nature enables the interpretation of the distinct Q-estimates as a function of the probability distribution, which in turn enables reasoning about the uncertainty in the mode, which is especially beneficial in the context of robotics. All the experiments conducted in [5] had a common goal location. Previous work done on goal-driven navigation includes the goal location in the visual feed, however, in the approach presented in [5], the location of the goal is not visible on the images. The robot is instead trained to identify the environmental landmarks only from photographic inputs. \par

The experimental setup in [5] compares the following 3 types of RL algorithms that learn from parallel experiences:
\subsection{Advantage-Actor-Critic (A2C)}
This algorithm is a synchronous implementation of A3C in which the same model is used to choose parallel actions. It is a policy-search method that aims to optimize a probability distribution over possible actions.
\subsection{n-step Q-learning}
This algorithm is a value-based method similar to the bootstrapped approach but one which maintains a single Q-function estimate.
\subsection{n-step bootstrapped Q-learning}
\textit{Bootstrapping} is the use of one or more estimated values in the update step for the similar kinds of estimated values. The performance of the bootstrapped Q-learning algorithm is significantly better, compared to the other 2 algorithms. 
Future work as described in [5] will be focusing on transfer to situations with agent and robot in a loop. Validation set performance provides evidence to show that closed-loop transfer is a feasible solution, although several issues are yet to be addressed.

\section{Conclusion}
This paper aims to serve as a reference guide for researchers looking to get into research in reinforcement learning applied to the field of robotics. The literature survey is at a fairly introductory level, aimed at aspiring researchers. Appropriately, we have covered the most essential concepts required for research in the field of reinforcement learning, with robotics in mind. Right from setting up of an experimental setup for real-world application to methods such as interactive replay, we also cover other concepts integral to the field of robotics, such as robots learning to solve tasks, learning synergies between prehensile and non-prehensile actions, and learning of locomotive behaviors in various terrains, all of which have been presented after much testing, in a well-thought-out experimental setup. \par

\end{document}